\documentclass[10pt,twocolumn,letterpaper]{article}

\usepackage[pagenumbers]{cvpr} 
\usepackage[accsupp]{axessibility}  
\usepackage{graphicx}
\usepackage{amsmath}
\usepackage{amssymb}
\usepackage{booktabs}

\usepackage[pagebackref,breaklinks,colorlinks]{hyperref}
\usepackage{xcolor}
\usepackage{makecell}

\usepackage[capitalize]{cleveref}
\crefname{section}{Sec.}{Secs.}
\Crefname{section}{Section}{Sections}
\Crefname{table}{Table}{Tables}
\crefname{table}{Tab.}{Tabs.}

\def\cvprPaperID{1501} 
\def\confName{CVPR}
\def\confYear{2022}

\newcommand{\modelname}{PTTR}

\begin{document}

\title{\modelname{}: 
Relational 3D Point Cloud Object Tracking with Transformer
}

\author{Changqing Zhou$^{1,3\dagger}$\thanks{Work done during an internship at Sensetime} \;
Zhipeng Luo$^{2,3}$\thanks{Equal contribution} \;
Yueru Luo$^{1\dagger}$ \;
Tianrui Liu$^{1,3}$ \; \\
Liang Pan$^{2}$ \thanks{Corresponding author} \;
Zhongang Cai$^{3}$ \;
Haiyu Zhao$^3$ \;
Shijian Lu$^1$ \; \\
$^1$ Nanyang Technological University 
$^2$ S-Lab, Nanyang Technological University 
$^3$ Sensetime Research \\
}
\maketitle

\begin{abstract}

In a point cloud sequence, 3D object tracking aims to predict the location and orientation of an object in the current search point cloud given a template point cloud.
Motivated by the success of transformers, 
we propose \textbf{P}oint \textbf{T}racking \textbf{TR}ansformer (\modelname{}), which efficiently predicts high-quality 3D tracking results in a coarse-to-fine manner with the help of transformer operations.
\modelname{} consists of three novel designs. 
\textbf{1)} Instead of random sampling, we design \textit{Relation-Aware Sampling} to preserve relevant points to given templates during subsampling. 
\textbf{2)} 
Furthermore, we propose a \textit{Point Relation Transformer} (PRT) consisting of a self-attention and a cross-attention module.
The global self-attention operation captures long-range dependencies to enhance encoded point features for the search area and the template, respectively.
Subsequently, we generate the coarse tracking results by matching the two sets of point features via cross-attention. 
\textbf{3)} Based on the coarse tracking results, we employ a novel \textit{Prediction Refinement Module} to obtain the final refined prediction.
In addition, we create a large-scale point cloud single object tracking benchmark based on the Waymo Open Dataset.
Extensive experiments show that \modelname{} achieves superior point cloud tracking in both accuracy and efficiency. 
Our code is available at \url{https://github.com/Jasonkks/PTTR}.
\end{abstract}


\section{Introduction} \label{intro}

With the rapid development of 3D sensors in the past decade, solving various vision problems \cite{qi2017pointnet,qi2017pointnet++,20203dssd,shi2019pointrcnn,luo2021unsupervised,xiao2021synlidar,xiao2022unsupervised,landrieu2018large,ren2022benchmarking,pan2021variational} with point clouds has attracted increasing attention due to the huge potential in applications such as autonomous driving, motion planning, and robotics. As a long-standing research problem in computer vision, object tracking with point clouds has also drawn wide research interests. 3D object tracking aims to detect not only object poses and positions in each frame but also object motion trajectories across consecutive frames. However, 3D tracking still faces a number of open and challenging problems such as LiDAR point cloud sparsity, random shape incompleteness, texture feature absence, etc.


\begin{figure}[t]
    \centering
    \includegraphics[width=1.0\linewidth]{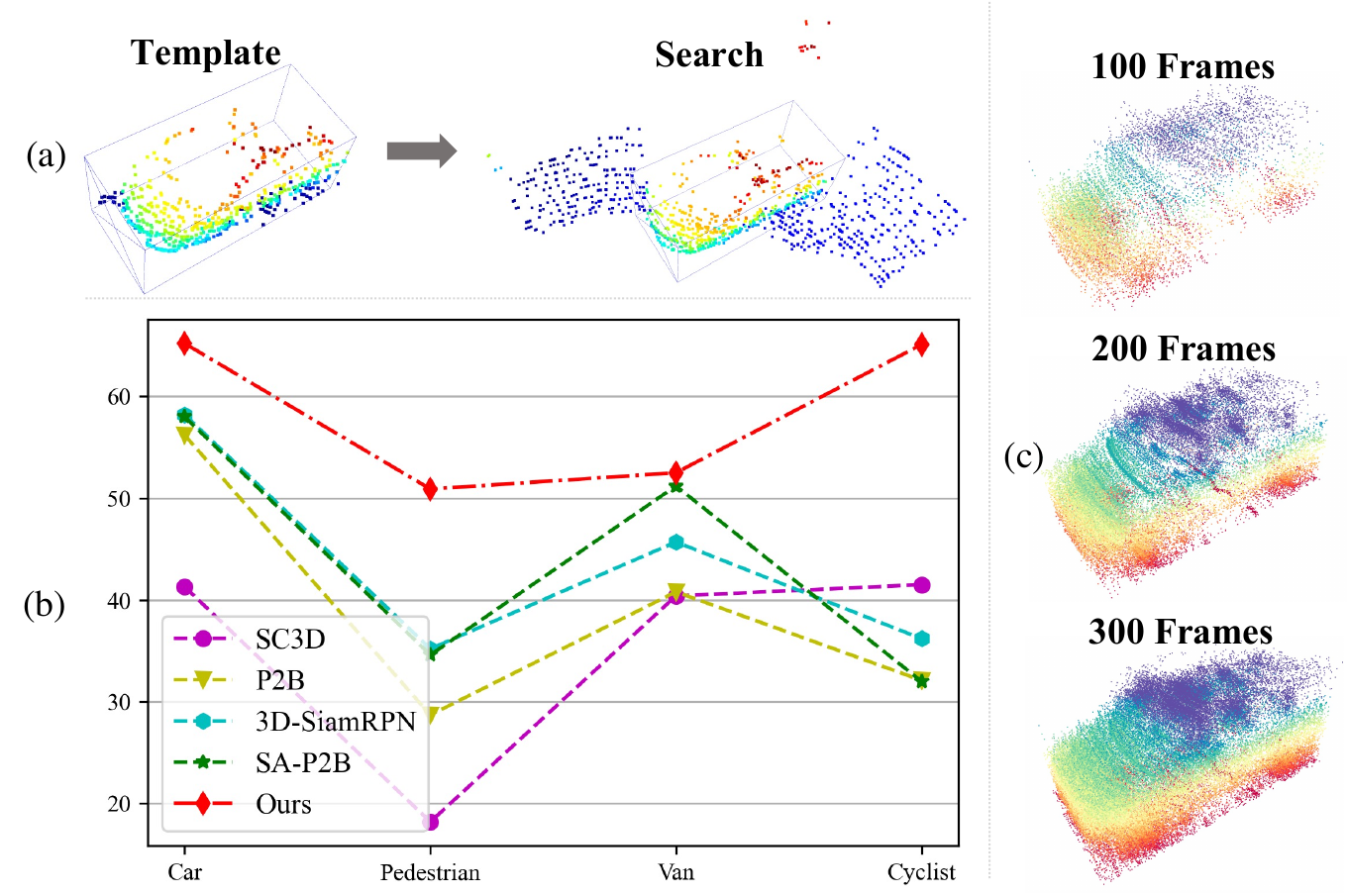}
    \vspace{-5mm}
    \caption{
    \textbf{(a)} 3D point cloud object tracking aims to track the target object based on a given template point cloud.
    \textbf{(b)} \modelname{} outperforms existing approaches by large margins on KITTI tracking dataset \cite{kitti}.
    \textbf{(c)} We visualize our tracking results over consecutive frames, including ``100 Frames'', ``200 Frames'' and ``300 Frames'', which
    demonstrate the robustness of \modelname{} for long-term tracking.}
    \label{fig:teaser_success}
    \vspace{-3mm}
\end{figure}

Existing 3D object tracking approaches can be largely categorized into two groups: multi-object tracking (MOT) and single-object tracking (SOT). 
MOT methods~\cite{weng20203d,wang2021joint,weng2020graph,yin2021center} generally adopt a detect-to-track strategy by first detecting objects in each frame and then matching the detections across consecutive frames based on the estimated location or speed.
In contrast, most SOT methods are only required to process a subset of point clouds, which usually come with much lower computational consumption and higher throughput.
We study SOT in this work, and our objective is to estimate the location and orientation of a single object in the search frame given an object template.

The pioneer 3D SOT method SC3D~\cite{2019sc3d} first generates a series of candidates given the last location of a specific object, and the prediction is made by selecting the best-matched candidate in the latent space.
However, it is not end-to-end trainable and suffers from low inference speed due to requiring a large number of candidates.
Without using many candidates, P2B~\cite{2020p2b} first use cosine similarity to fuse features of the search region with the template and then adopts the prediction head of VoteNet~\cite{qi2019deep} to generate the final prediction. 
Following P2B, SA-P2B~\cite{zhou2021structure} adds an extra auxiliary network to predict the object structure. 
In a similar framework, 3D-SiamRPN~\cite{fang20203d} uses a cross-correlation module for feature matching and an RPN head for final prediction.
These methods~\cite{2019sc3d, 2020p2b, zhou2021structure} essentially perform a linear matching process between features in the search domain and the template, which 
cannot adapt to different 3D observations caused by random noise, sparsity, and occlusions.
Moreover, the inclusion of complex prediction heads as in detection models highly limits their tracking speed, which is a crucial factor for online applications.

In this work, we design Point Tracking TRansformer (\modelname{}), a novel tracking paradigm that achieves high-quality 3D object tracking in a coarse-to-fine fashion.
Specifically, \modelname{} first extracts point features from the template and search area individually using the PointNet++ \cite{qi2017pointnet++} backbone. 
To alleviate the point sparsity issue, we propose a sampling strategy termed \textit{Relation-Aware Sampling}, which can preserve more points that are relevant to the given template by leveraging the relation-aware feature similarities between the search and the template. 
We then propose a novel \textit{Point Relation Transformer} (PRT) equipped with \textit{Relation Attention Module} to match search and template features and generate a coarse prediction
based on the matched feature. PRT first utilizes a self-attention operation to adaptively aggregate point features for the template and the search area individually, and then performs feature matching with a cross-attention operation.  
Moreover, we propose a lightweight \textit{Prediction Refinement Module} to refine the coarse prediction with local feature pooling. We highlight that \modelname{} is more efficient than existing methods despite its prediction refinement process.

KITTI tracking dataset \cite{kitti} has been widely adopted in 3D tracking evaluations. However, it has clear constraints including limited sample size and highly imbalanced class distributions. We create a new point cloud tracking benchmark named Waymo SOT Dataset based on Waymo Open Dataset \cite{waymo}, which has a large sample size as well as balanced class distributions. The new benchmark is thus complementary to the KITTI tracking dataset by offering more holistic and comprehensive evaluations to the 3D tracking research community. 
Extensive experiments on both datasets demonstrate the superior performance of \modelname{} in both accuracy and efficiency.

Our \textbf{key} contributions are summarized as:
\textbf{1)} We propose \modelname{}, a transformer-based 3D point cloud object tracking method, which employs a novel coarse-to-fine tracking paradigm to first generate coarse global prediction and refine it with Local Pooling. 
\textbf{2)} We design two novel modules in \modelname{} including Point Relation Transformer for effective feature aggregation and matching, and Relation-Aware Sampling for preserving more template-relevant points.
\textbf{3)} \modelname{} surpasses previous SoTA methods by large margins in performance with lower computational cost.
\textbf{4)} We generate a new large-scale point cloud tracking dataset based on the Waymo Open Dataset \cite{waymo} to facilitate more comprehensive evaluations of 3D object tracking approaches.

\section{Related Works}

\begin{figure*}[t]
    \centering
    \includegraphics[width=1.0\linewidth]{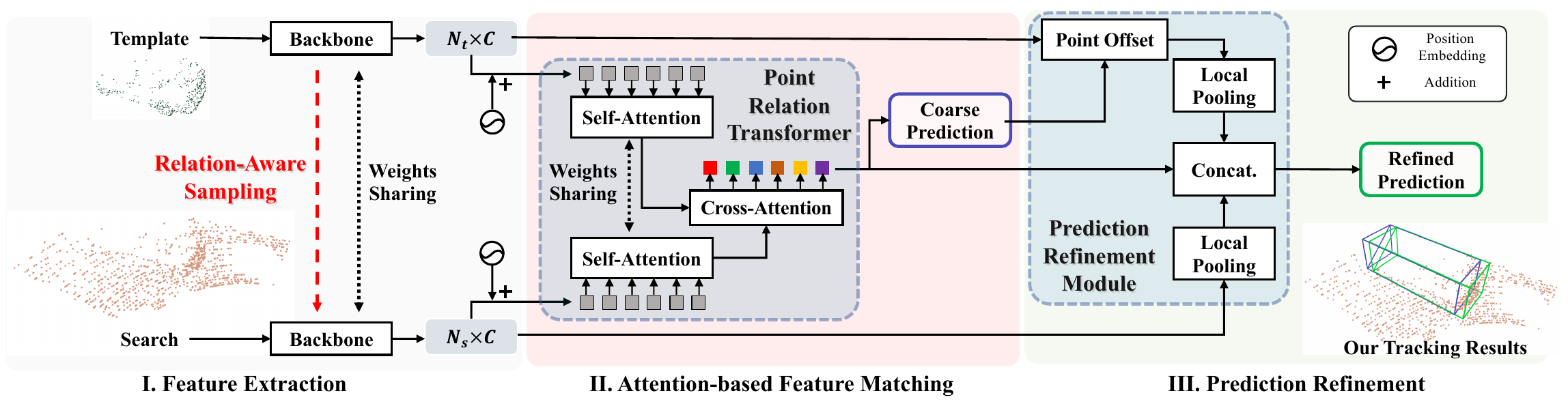}
    \vspace{-7mm}
    \caption{\textbf{Overview of our proposed \modelname{}.} The network mainly consists of three parts: feature extraction, attention-based feature matching and prediction refinement. The backbone is used to extract features from input point clouds. We modify PointNet++ \cite{qi2017pointnet++} with our proposed \textit{Relation-Aware Sampling} to help select more positive points from the search area. For feature matching, we propose \textit{Point Relation Transformer} equipped with \textit{Relation Attention Module} to match search and template features. In the prediction stage, we propose a \textit{Prediction Refinement Module} to generate predictions in a coarse-to-fine manner.}
    \label{fig:overview}
    \vspace{-3mm}
\end{figure*}

\noindent\textbf{2D Object Tracking.} Majority of the recent 2D object tracking methods follow the Siamese network paradigm, composed of two CNN branches with shared parameters that help project inputs into the same feature space. \cite{tao2016siamese} employs Siamese network to learn a generic matching function for different objects. At inference, a bunch of candidates are used to match the original target and the one that matches best is chosen as the prediction. \cite{bertinetto2016fully} proposes a fully-convolutional Siamese architecture to locate the target object in a larger search area. 
\cite{guo2017learning} proposes a dynamic Siamese network that learns to transform the target appearance and suppress the background. \cite{li2018high,li2019siamrpn++} apply Siamese network to extract features and use pair-wise correlation separately for classification branch and regression branch of the region proposal network (RPN). 2D tracking approaches are not directly applicable to point clouds as they are driven by 2D CNN architectures and they are not designed to address the unique challenges of 3D tracking. 

\noindent\textbf{3D Object Tracking.} 3D object tracking can be roughly divided into two categories: multi-object tracking (MOT) and single-object tracking (SOT). Most MOT approaches adopt a detect-to-track strategy and mainly focus on data association \cite{xiang2015learning,karunasekera2019multiple}. \cite{weng2019baseline} first proposes a 3D detection module to provide the 3D bounding boxes, then use 3D Kalman Filter to predict current estimation, and match them using Hungarian algorithm. \cite{wang2021joint} proposes to use GNN to model relationships among different objects both spatially and temporally, while \cite{yin2021center} uses a closest distance matching after speed compensation.
SOT methods focus on tracking a single object given a template. SC3D \cite{2019sc3d} proposes to match feature distance between candidates and target and regularize the training using shape completion. P2B \cite{2020p2b} matches search and template features with cosine similarity and employs Hough Voting \cite{qi2019deep} to predict the current location. SA-P2B \cite{zhou2021structure} proposes to learn the object structure as an auxiliary task. 3D-SiamRPN \cite{fang20203d} uses a RPN \cite{ren2015faster} head to predict the final results. BAT \cite{zheng2021box} encodes box information in Box Cloud to incorporate structural information. MLVSNet \cite{wang2021mlvsnet} proposes to perform multi-level Hough voting for aggregating information from different levels. PTT \cite{shan2021ptt} proposes a Point-Track-Transformer module to weight features' importance.  Most existing SOT methods either use cosine similarity or cross-correlation to match the search and template features, which are essentially linear matching processes and cannot adapt to complex situations where random noise and occlusions are involved. Moreover, the use of detection model prediction heads leads to high computation overheads. Our proposed method in this paper address the above limitations.

\noindent\textbf{Vision Transformers.} Transformer \cite{vaswani2017attention} was first proposed as an attention-based building block in machine translation to replace the RNN architecture. Recently, a number of works \cite{liu2021swin,chu2021twins,carion2020end,zhu2020deformable,xie2021segformer,zhang2021meta,zhang2022accelerating} apply transformer on 2D vision tasks and achieve great success. Most of these attempts divide the images into overlapping patches and then regard each patch as a token to further apply the transformer architecture.
In the 3D domain, PCT \cite{guo2021pct} generates positional embedding using 3D coordinates of points and adopts transformer with an offset attention module to enrich features of points from its local neighborhood. Point Transformer \cite{zhao2021point} adopted vectorized self-attention network \cite{zhao2020exploring} for local neighbours and designed a Point Transformer layer that is order-invariant to suit point cloud processing. \cite{engel2021point} proposes SortNet to gather spatial information from point clouds, which sorts the points by learned scores to achieve order invariance. All of these works focus on shape classification or part segmentation tasks.
The attention mechanism in transformer offers correlation modeling with global receptive field, which makes it a good candidate for the 3D tracking problem where feature matching is required. We propose a novel transformer-based module to utilize the attention mechanism for feature aggregation and matching.

\section{Method} \label{method}

\subsection{System Overview} \label{sec:overview}
Given a 3D point cloud sequence, 3D object tracking aims to estimate the object location and orientation in each point cloud observation, \ie the search point cloud $P^s \in \mathbb{R}^{N_s \times 3}$, by predicting a bounding box conditioned on a template point cloud $P^t \in \mathbb{R}^{N_t \times 3}$. 
To this end, we propose \modelname{}, a novel coarse-to-fine framework for 3D object tracking.
As shown in Fig.~\ref{fig:overview}, \modelname{} performs 3D point cloud tracking with three main stages: 1) Feature Extraction (Sec.~\ref{sec:sample}); 2) Attention-based Feature Matching (Sec.~\ref{sec:transformer}); and 3) Prediction Refinement (Sec.~\ref{sec:refinement}).

\noindent\textbf{Feature Extraction.}
Following previous methods \cite{2020p2b, 2019sc3d, zhou2021structure, fang20203d}, we employ PointNet++~\cite{qi2017pointnet++} as the backbone to extract multi-scale point features from the template and the search.
However, important information loss may occur during random subsampling in the original PointNet++.
We therefore propose a novel 
\textit{Relation-Aware Sampling} 
to preserve more points relevant to the given template by leveraging relation-aware feature similarities.

\noindent\textbf{Attention-based Feature Matching.}
Different from previous methods that often use cosine similarity \cite{2019sc3d,2020p2b,zhou2021structure} or linear correlation \cite{fang20203d} for matching the template and the search, 
we utilize novel attention operations and propose \textit{Point Relation Transformer} (PRT).
PRT first utilizes a self-attention operation to adaptively aggregate point features for the template and the search area individually, and then performs feature matching with cross-attention.
The coarse prediction is generated based on the output of PRT.

\noindent\textbf{Prediction Refinement.}
The coarse prediction is further refined with a lightweight \textit{Prediction Refinement Module} (PRM), which results in a coarse-to-fine tracking framework.
Based on the coarse predictions, we first conduct a Point Offset operation for seed points from the search to estimate their corresponding seed points in the template.
Afterwards, we employ a Local Pooling operation for the seed points from both point clouds respectively, and then concatenate the pooled features with the matched features from PRT for estimating our final prediction.


\begin{figure}[t]
    \centering
    \includegraphics[width=1.0\linewidth]{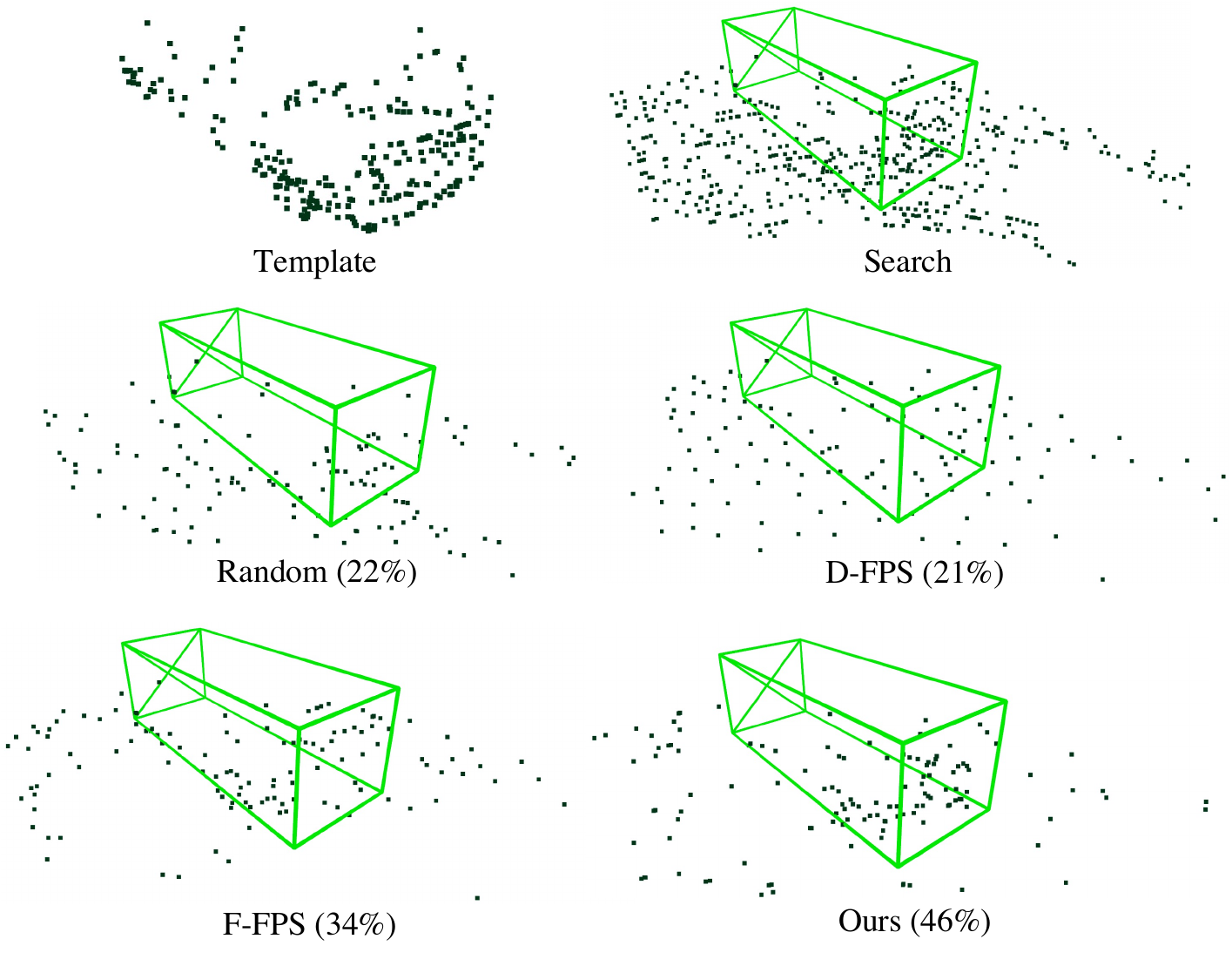}
    \vspace{-8mm}
    \caption{\textbf{Comparison of sampling methods}. We show the sampled points from the search area using different sampling methods. Our proposed sampling method preserves the most number of points belonging to the object. The percentage in the figure represents the ratio of positive points to all sampled points.}
    \label{fig:sample_method}
\vspace{-3mm}
\end{figure}

\subsection{Relation-Aware Feature Extraction} \label{sec:sample}

As one of the most successful backbones, PointNet++~\cite{qi2017pointnet++} introduces a hierarchical architecture with multiple distance-farthest point sampling (D-FPS) and ball query operations, which effectively exploits multi-scale point features.
Most existing 3D object tracking methods~\cite{2020p2b, zhou2021structure, fang20203d} use PointNet++ for feature extraction.
However, it has a non-negligible disadvantage for object tracking: the D-FPS sampling strategy used in PointNet++ tends to generate random samples that are uniformly-distributed in the euclidean space, which often leads to important information loss during the sampling process.
In particular, the search point cloud often has a much larger size than the template, and therefore D-FPS sampling inevitably keeps a substantial portion of background points and leads to sparse point distribution for the object of interest, which further challenges the subsequent template searching using feature matching.
To alleviate this problem, previous methods use either random point sampling~\cite{2020p2b, zhou2021structure} or feature-farthest point sampling (F-FPS)~\cite{20203dssd}.
However, the problem of substantial foreground information loss during sampling is not fully resolved.

\smallskip
\noindent\textbf{Relation-Aware Sampling.}
In contrast, we propose to use a novel sampling method dubbed \textit{Relation-Aware Sampling} (RAS) to preserve more points relevant to the given template by considering relational semantics.
Our key insight is that the region of interest in the search point cloud should have similar semantics with the template.
Therefore, points in search area with higher semantic feature similarities to the template points are more likely to be foreground points.
Specifically, given the template point features $\mathbf{X}^t \in \mathbb{R}^{N_t \times C}$ and search area point features $\mathbf{X}^s \in \mathbb{R}^{N_s \times C}$, we first calculate the pairwise point feature distance matrix $\mathbf{D} \in \mathbb{R}^{N_s \times N_t}$:
\begin{equation}
    \mathbf{D}_{ij} = ||\mathbf{x}^s_i - \mathbf{x}^t_j||_2,\;\;  \forall\, \mathbf{x}^s_i \in \mathbf{X}^s,\;\; \forall\, \mathbf{x}^t_j \in \mathbf{X}^t,
\end{equation}
where $||\cdot||_2$ denotes L2-norm, and $N_s$ and $N_t$ are the current number of points from the search area and the template, respectively. 
Afterwards, we compute the minimum distance $\mathbf{V} \in \mathbb{R}^{N_s}$ by considering the distance between each point from search and its nearest point from template in the feature space:
\begin{equation}
    \mathbf{V}_i = \min_{j=1}^{N_t} (\mathbf{D}_{ij}),\;\;  \forall\, i \in \{1,2,...,N_s\}.
\end{equation}

Following previous methods~\cite{2020p2b, 2019sc3d, zhou2021structure, fang20203d},
we update the template point cloud for each frame by using the tracking result from the previous search point cloud.
In the case that low-quality tracking predictions are encountered in difficult situations, the newly formed templates might mislead the RAS and lead to unfavorable sampling results. Moreover, the inclusion of background information offers useful contextual information for the localization of the tracked object. To improve the robustness of the sampling process, we adopt a similar strategy as in \cite{20203dssd} to combine our proposed RAS with randomly sampling. In practice, we sample half of the points with RAS, while the rest of the points are obtained via random sampling. We show the effects of different sampling approaches in Fig.~\ref{fig:sample_method}. It can be observed that the proposed sampling method can preserve the most object points.

\subsection{Relation-Enhanced Feature Matching} \label{sec:transformer}
Existing 3D object tracking methods perform feature matching between the search point cloud and template by using cosine similarity \cite{2019sc3d,2020p2b,zhou2021structure} or linear correlation \cite{fang20203d}.
On the other hand, motivated by the success of various attention-based operations for computer vision applications~\cite{pan2021variational,zhao2021point,vaswani2017attention,guo2021pct}, we strive to explore attention-based mechanism for 3D tracking, which can adapt to different noisy point cloud observations. Although PTT \cite{shan2021ptt} utilizes transformer in their model, they still match template and search point cloud by cosine similarity and the transformer module is only used for feature enhancement.

\begin{figure}[t]
    \centering
    \includegraphics[width=0.6\linewidth]{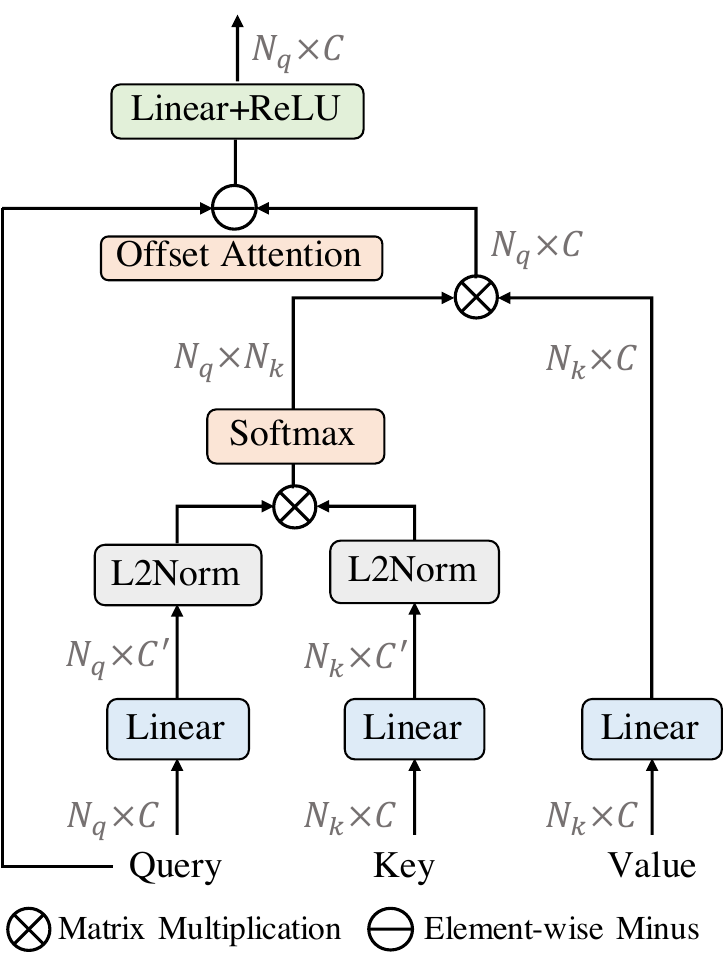}
    \vspace{-3mm}
    \caption{\textbf{Architecture of the Relation Attention Module (RAM).} 
    RAM first projects query, key and value into a latent feature space and then estimates the attention matrix by multiplying the L2-normalized query and key features. The attention matrix is then applied on the value feature to obtain the attention product before the offset attention~\cite{guo2021pct} operation and Linear and ReLU layers for injection of non-linearity.
    }
    \label{fig:relation attention}
\vspace{-3mm}
\end{figure}

\smallskip
\noindent\textbf{Relation Attention Module.}
Inspired by recent works studying feature matching~\cite{wang2021transformer, wang2018cosface,wang2017normface,chen2020simple}, we propose the \textit{Relation Attention Module} (RAM) (shown in Fig.~\ref{fig:relation attention}) to adaptively aggregate features by predicted attention weights.
Firstly, RAM employs linear projection layers to transform the input feature vectors ``Query'', ``Key'' and ``Value''.
Instead of naively calculating the dot products between ``Query'' and ``Key'', RAM predicts the attention map by calculating the cosine distances between the two sets of L2-normalized feature vectors.
With the help of L2-normalization, RAM can prevent the dominance of a few feature channels with extremely large magnitudes. 
Subsequently, the attention map is normalized with a Softmax operation.
In order to sharpen the attention weights and meanwhile reduce the influence of noise~\cite{guo2021pct}, we employ the offset attention to predict the final attention map by subtracting the query features with the previously normalized attention map.
Consequently, the proposed RAM can be formulated as:
\begin{align}
    \text{Attn}(\mathbf{Q}, \mathbf{K}, \mathbf{V}) = \phi\big(\mathbf{Q}& - \text{softmax}(\mathbf{A})\cdot(W_v \mathbf{V})\big),
\end{align}
where $\phi$ represents the linear layer and ReLU operation applied to the output features, the attention matrix $\mathbf{A} \in \mathbb{R}^{N_q \times N_{k}}$ is obtained by:
\begin{align}
    \mathbf{A} = \overline{\mathbf{Q}} \cdot \overline{\mathbf{K}}^{^\top},\;
    \overline{\mathbf{Q}} = \frac{W_q \mathbf{Q}}{||W_q \mathbf{Q}||_2}, \; 
    \overline{\mathbf{K}} = \frac{W_k \mathbf{K}}{||W_k \mathbf{K}||_2},
\end{align}
where $\|\cdot\|_2$ is the L2-norm,
$\mathbf{Q}, \mathbf{K}, \mathbf{V}$ represent the input ``Query'', ``Key'' and ``Value'' respectively, and $W_q$, $W_k$ and $W_v$ denote the corresponding linear projections.

\smallskip
\noindent\textbf{Point Relation Transformer.}
By incorporating RAM,
we propose the \textit{Point Relation Transformer} (PRT) module
to adaptively exploit the correlations between point features for context enhancement.
The PRT module firstly performs a self-attention operation for the search and the template features, respectively. 
Subsequently, PRT employs a cross-attention operation for gathering cross-contextual information between the two point clouds.
Both operations use global attentions, where all input point feature vectors are considered as tokens.
Formally, PRT is formulated as:
\begin{align}
    \bar{\mathbf{X}}^s & = \text{Attn}(\mathbf{X}^s), \;\;\text{and}\;\;
    \bar{\mathbf{X}}^t  = \text{Attn}(\mathbf{X}^t), \\
    \hat{\mathbf{X}}^s & = \text{Attn}(\bar{\mathbf{X}}^s, \bar{\mathbf{X}}^t, \bar{\mathbf{X}}^t), 
\end{align}
where $\text{Attn}(\mathbf{Q}, \mathbf{K}, \mathbf{V})$ denotes our proposed \textit{Relation Attention Module}, $\hat{\mathbf{X}}^s$ denotes the matched features, and $\bar{\mathbf{X}}^s$ and $\bar{\mathbf{X}}^t$ denote the enhanced search and template features respectively.
By using global self-attention, the exploited features can obtain a global understanding for the current observation.
Note that the self-attention use the same point features as $\mathbf{Q}, \mathbf{K}, \mathbf{V}$, and both self-attention operations share weights so as to project the search and template features into the same latent space.
Thereafter, the cross-attention performs pairwise matching between query tokens $\bar{\mathbf{X}}^s$ and key tokens $\bar{\mathbf{X}}^t$, which exploits cross-contextual information for $\hat{\mathbf{X}}^s$ by capturing correlations between the two sets of point features.
Based on the relation enhanced point features $\hat{\mathbf{X}}^s$, we can generate our coarse prediction results for 3D object tracking.

\subsection{Coarse-to-Fine Tracking Prediction} \label{sec:refinement}
Majority of the existing point tracking approaches adopt prediction heads of detection models to generate the predictions, e.g. P2B \cite{2020p2b} adopts the clustering and voting operations of VoteNet \cite{qi2019deep} and 3D-SiamRPN \cite{fang20203d} uses a RPN \cite{ren2015faster,li2018high} head. 
However, these prediction heads introduce extra computation overheads, which largely limits their efficiency.
To circumvent this issue, 
we propose a novel coarse-to-fine tracking framework.
The coarse prediction $\mathbf{Y}^c$ is predicted by directly regressing the relation enhanced features $\hat{\mathbf{X}}^s$ from the proposed PRT module with Multi-Layer-Perceptron (MLP).
Remarkably, $\mathbf{Y}^c$ provides faithful tracking predictions for most cases, and also surpasses the tracking performance of SoTA methods.

\smallskip
\noindent\textbf{Prediction Refinement Module.}
To further refine the tracking predictions, we propose a lightweight \textit{Prediction Refinement Module} (PRM) to predict our final predictions $\mathbf{Y}^f$ based on $\mathbf{Y}^c$.
Specifically, we use the sampled points from the search point cloud as seed points, and then we estimate their correspondences in the template by using an offset operation for $\mathbf{Y}^c$. 
Afterwards, we encode local discriminative feature descriptors for the seed points from both sources, which is achieved by using Local Pooling operations for grouped neighboring point features.
The neighboring features are grouped by using ball-query operations with a fixed radius $r$. 
In the end, we concatenate $\hat{\mathbf{X}}^s$ with the pooled features from the source and the target, based on which we generate the final prediction $\mathbf{Y}^f$:  
\begin{align}
    \mathbf{Y}^f = \gamma([\mathbf{F}^s, \mathbf{F}^t, \hat{\mathbf{X}}^s]),
\end{align}
where $\mathbf{F}^s$ and $\mathbf{F}^t$ are the pooled features from search and template respectively, $[\cdot]$ denotes concatenation operation, and $\gamma$ represents the MLP networks. 
We highlight that even with the refinement stage, our proposed method still has lower computational complexity than existing methods thanks to the lightweight design. 

\smallskip
\noindent\textbf{Training Loss.}
Our \modelname{} is trained in an end-to-end manner.
The coarse prediction $\mathbf{Y}^c$ and the final prediction $\mathbf{Y}^f$ are in the same form that each contains a classification component $\mathbf{Y}_{cls} \in \mathbb{R}^{N_s\times1}$ and a regression component $\mathbf{Y}_{reg} \in \mathbb{R}^{N_s\times4}$, where $N_s$ denotes the number of sampled points from the search area. $\mathbf{Y}_{cls}$ predicts the objectiveness of each point, and $\mathbf{Y}_{reg}$ consists of the predicted offsets along each axis $\{\Delta{x}, \Delta{y}, \Delta{z}\}$ with an additional rotation angle offset $\Delta{\theta}$. 
For each prediction, we use a classification loss $\mathcal{L}_{cls}$ defined by binary cross entropy, and a regression loss $\mathcal{L}_{reg}$ computed by mean square error.
Consequently, our overall loss function is formulated as:
\begin{align}
    \mathcal{L}_{total} =& \mathcal{L}_{cls}(\mathbf{Y}^c_{cls}, \mathbf{Y}^{gt}_{cls}) + \mathcal{L}_{reg}(\mathbf{Y}^c_{reg}, \mathbf{Y}^{gt}_{reg}) +\\ &\lambda\big(\mathcal{L}_{cls}(\mathbf{Y}^f_{cls}, \mathbf{Y}^{gt}_{cls}) + \mathcal{L}_{reg}(\mathbf{Y}^f_{reg}, \mathbf{Y}^{gt}_{reg})\big),
\end{align}
where $\mathbf{Y}^{gt}_{(\cdot)}$ denotes the corresponding ground truth, and $\lambda$ is a weighting parameter.

\begin {table}[t]
\setlength{\tabcolsep}{15pt}
\small
\caption {\textbf{Dataset Statistics.} Number of \textit{tracklets / samples} from different categories in KITTI \cite{kitti} and Waymo SOT Dataset.} \label{tab:dataset} 
\vspace{-6mm}
\begin{center}\setlength{\tabcolsep}{1pt}{
\scalebox{1.0}{
\begin{tabular}{l|c|c|c|c}
\Xhline{1pt}
{\bf Dataset} & {\bf Car / Vehicle } & {\bf Pedestrian} & {\bf Van} & {\bf Cyclist} \\ 
\hline\hline
KITTI (train) & \scriptsize 441 / 19522 & \scriptsize 96 / 4600 & \scriptsize 38 / 1994 & \scriptsize 27 / 1529 \\
KITTI (test) & \scriptsize 120 / 6424 & \scriptsize 62 / 6088 & \scriptsize 16 / 1248 & \scriptsize 8 / 308 \\
\hline
Waymo (train) & \scriptsize 16119 / 241544 & \scriptsize \;15452 / 249800\; & \scriptsize - & \scriptsize \;1066 / 22389 \\
Waymo (test) & \scriptsize 1658 / 53377 & \scriptsize 949 / 27308 & \scriptsize - & \scriptsize 138 / 5374 \\
\Xhline{1pt}
\end {tabular}}
\label{tab:dataset}
}
\end{center}
\vspace{-4mm}
\end {table}

\section{Waymo 3D Single Object Tracking Dataset}
\noindent\textbf{Existing Benchmark.}
Existing methods carry out their evaluations on the KITTI \cite{kitti} tracking dataset. 
The data split, tracklet generation, and evaluation metrics are specified in \cite{2019sc3d}. 
However, we find this dataset offers a limited number of samples while the object classes are highly imbalanced. 
Tab.~\ref{tab:dataset} shows the statistics of the datasets and we can observe that the car category accounts for the majority of the tracklets and samples, while some classes have few examples. 
\cite{2020p2b} studies the impact of the limited training samples and finds the performance severely degrades with insufficient training data. 
Therefore, we believe it will be beneficial 
if a large-scale dataset is available. 

\smallskip
\noindent\textbf{Constructing Waymo SOT Dataset.}
Fortunately, we find that the recently released Waymo Open Dataset \cite{waymo} is able to fulfill this need. 
Although Waymo does not directly contain a SOT dataset, in its detection dataset, every object is not only annotated with the bounding box but also with a unique object ID, which makes it feasible to extract tracklets from the point cloud sequences. 
To alleviate the class imbalance issue, we use $10\%$ of the training and validation sequences to produce Vehicle tracklets, $20\%$ of the sequences for Pedestrian, and all the sequences for Cyclist since it is the rarest. 
We eliminate objects with less than 10 points inside the ground truth bounding box and remove tracklets with lengths less than 3 frames. 

\smallskip
\noindent\textbf{Properties and Advantages.}
The statistics of the generated dataset are reported in Tab.~\ref{tab:dataset}. 
Although the Waymo dataset does not differentiate different vehicles, such as cars and vans, the Waymo SOT dataset is of a significantly larger scale with a more balanced class distribution than the KITTI tracking dataset. 
For example, there are 15,452 ``Pedestrian'' tracklets in the proposed Waymo SOT Dataset, while KITTI Tracking dataset only has 96 tracklets.


\section{Experiments}

We evaluate our proposed method on both KITTI \cite{kitti} and Waymo SOT Dataset for comprehensive comparison.

\smallskip
\noindent\textbf{Evaluation Metrics.}
Following the evaluation metrics of P2B \cite{2020p2b} and measure the ``Success'' and ``Precision''.
Specifically, ``Success'' is defined as the IoU between predicted boxes and the ground truth, and ``Precision'' measures the AUC (Area Under Curve) of the distance between prediction and ground truth box centers from 0 to 2 meters.

\smallskip
\noindent\textbf{Template and Search.} 
During training, we use the ground truth bounding box to crop the point cloud to form the template. 
In order to simulate the disturbances the model might encounter, we add random distortions to augment the bounding boxes with a range of [-0.3 to 0.3] along x, y, z axes. 
For both training and testing, we extend the box with a ratio of 0.1 to include some background points. 
We enlarge the template bounding box by 2 meters in all directions to form the search area.

\smallskip
\noindent\textbf{Model Details.} 
We use PointNet++\cite{qi2017pointnet++} with 3 set-abstraction layers as the backbone. 
The radius of these SA layers is set to 0.3, 0.5, 0.7 meters, respectively.
In the first stage, we use a 3-layer MLP for classification and regression, respectively. 
Each layer is followed by a BN \cite{ioffe2015batch} layer and an ReLU \cite{agarap2018deep} activation layer. 
In PRM, the Local Pooling is conducted with a ball-query operation and a grouping operation \cite{qi2017pointnet++} with a radius of 1.0 meter.
After pooling, we obtain the concatenated features that are fed into a 5-layer MLP for generating the final predictions.

\smallskip
\noindent\textbf{Training and Testing.} 
For KITTI tracking dataset, we train the model for 160 epochs with a batch size of 64. We use Adam optimizer \cite{kingma2014adam} with an initial learning rate of 0.001 and reduce it by 5 every 40 epochs. 
For the Waymo SOT Dataset, we train the model for 80 epochs with the same initial learning rate and reduce the learning rate every 20 epochs. 
During testing, we use the previous prediction result as the next template. 
In line with \cite{2019sc3d,2020p2b}, we use the ground truth bounding box as the first template.

\subsection{3D Tracking on KITTI Tracking Dataset}

\begin{figure}[t]
    \centering
    \includegraphics[width=1.0\linewidth]{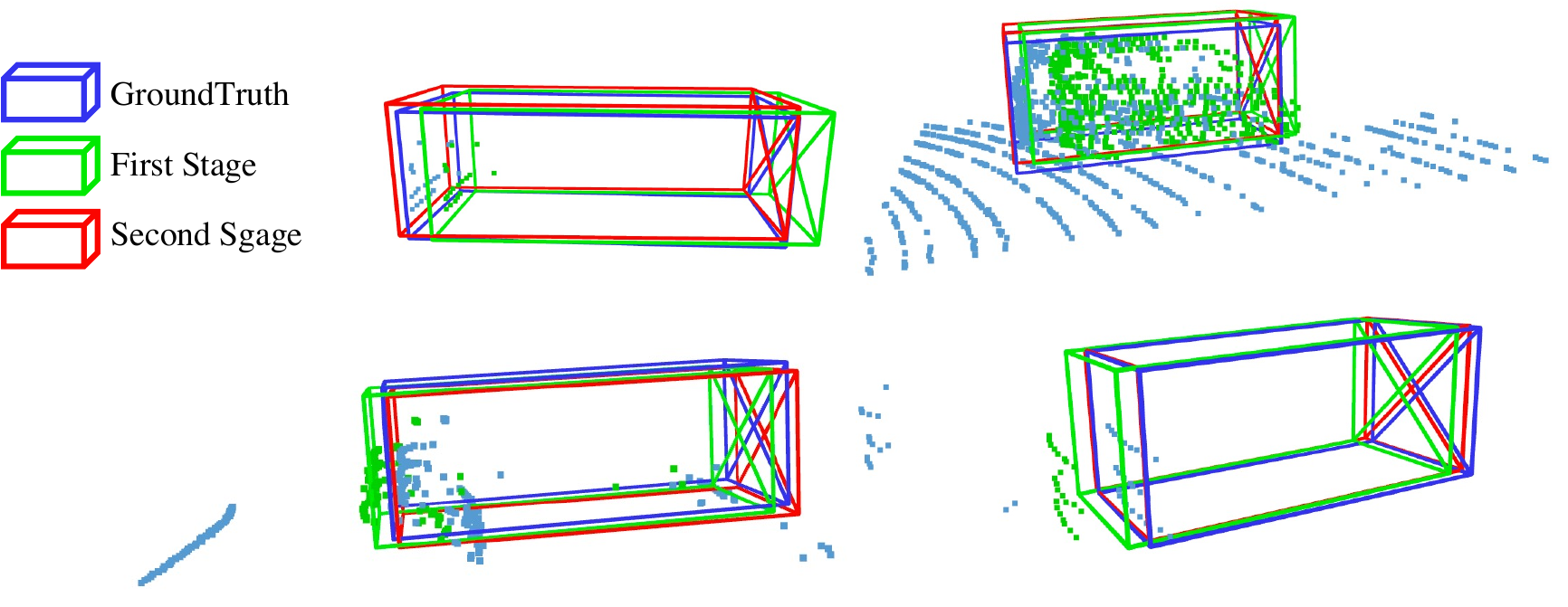}
    \vspace{-7mm}
    \caption{\textbf{Visualization of prediction refinement.}  
    We present 4 different objects to show that the refined stage further corrects the coarse predictions through local feature pooling.}
    \label{fig:second stage}
    \vspace{-3mm}
\end{figure}

\begin{figure*}[t]
    \centering
    \includegraphics[width=1.0\linewidth]{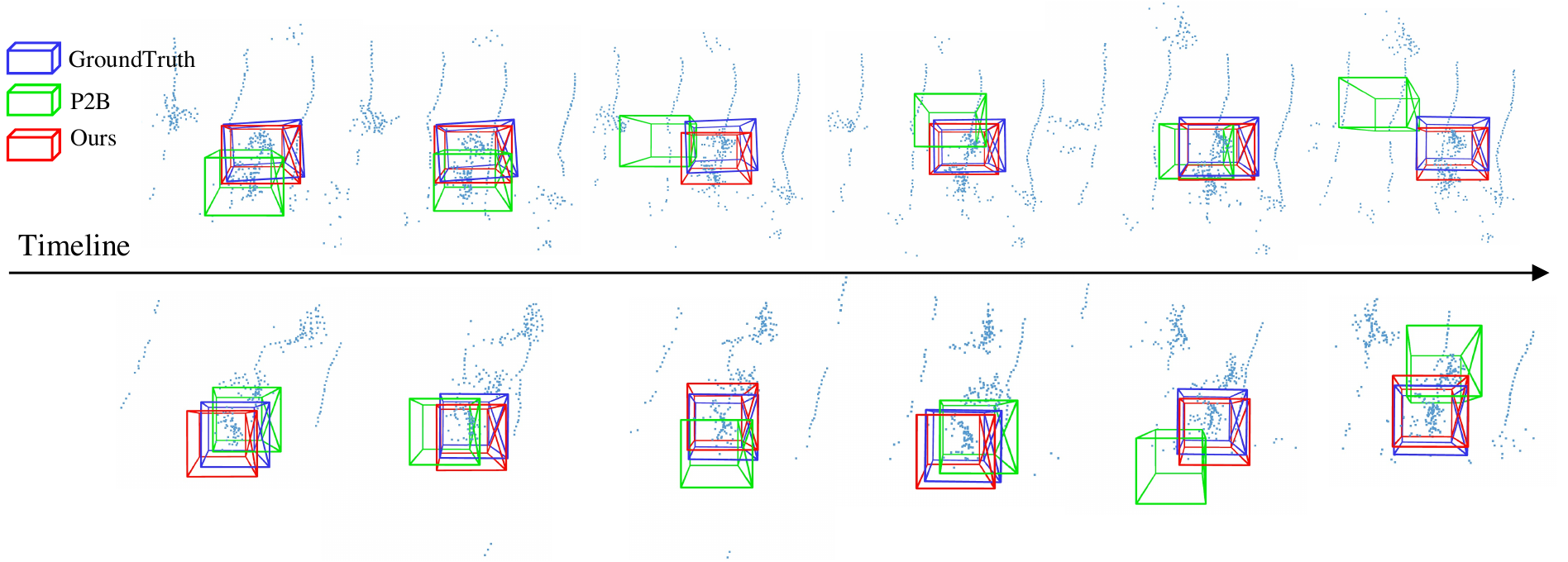}
    \vspace{-12mm}
    \caption{\textbf{Qualitative comparison with P2B \cite{2020p2b}.} We compare \modelname{} with P2B over two pedestrian tracking sequences. It can be observed that P2B is prone to matching mistakes when multiple instances are close, while \modelname{} generates more reliable predictions.}
    \label{fig:ours vs p2b}
\end{figure*}

To achieve fair comparisons against previous methods, we follow the data split and processing specified in \cite{2019sc3d,2020p2b}.
As reported in Tab.~\ref{tab:kitti_experiment}, \modelname{} surpasses the previous state-of-the-art method by a significant margin of 8.4 and 10.4 in terms of average Success and Precision. 
In particular, \modelname{} significantly outperforms SA-P2B on the challenging categories, Pedestrian and Cyclist.
As shown in Fig.~\ref{fig:ours vs p2b}, we compare the proposed \modelname{} against P2B \cite{2020p2b} over two pedestrian point cloud sequences. 
P2B often makes wrong predictions when multiple instances are close, while \modelname{} is able to generate stable and reliable predictions.
The impressive performance gains are largely attributed to the proposed PRT and RAS modules, which will be further explained in the ablation studies in Sec.~\ref{sec:ablations}.
Moreover, we visualize the coarse and refined predictions in Fig.~\ref{fig:second stage}.
The coarse predictions are further corrected in the refinement stage, especially when point sparsity or large movements are present. 
It demonstrates that our proposed prediction refinement via local feature pooling is able to adapt to challenging situations and generate robust predictions. 

\begin{table}[t]
\setlength{\tabcolsep}{20pt}
\small
\caption {\textbf{Performance comparison on the KITTI dataset.} Success / Precision are used for evaluation.} \label{tab:kitti_experiment} 
\vspace{-6mm}
\begin{center}\setlength{\tabcolsep}{1pt}{
\scalebox{1.0}{
\begin{tabular}{l|c|c|c|c|c}
\Xhline{1pt}
{\bf \fontsize{6.75}{6.75}\selectfont Method} & {\bf \fontsize{6.75}{6.75}\selectfont Car } & {\bf \fontsize{6.75}{6.75}\selectfont Pedestrian} & {\bf \fontsize{6.75}{6.75}\selectfont Van} & {\bf \fontsize{6.75}{6.75}\selectfont Cyclist} & \bf \fontsize{6.75}{6.75}\selectfont Average\\ 
\hline\hline
\fontsize{6.5}{6.5}\selectfont SC3D \cite{2019sc3d} & \scriptsize \;41.3 / 57.9\; & \scriptsize 18.2 / 37.8 & \scriptsize \;40.4 / 47.0\; & \scriptsize \;41.5 / 70.4\; & \scriptsize \;35.4 / 53.3\; \\
\fontsize{6.5}{6.5}\selectfont P2B \cite{2020p2b} & \scriptsize 56.2 / 72.8 & \scriptsize 28.7 / 49.6 & \scriptsize 40.8 / 48.4 & \scriptsize 32.1 / 44.7 & \scriptsize 39.5 / 53.9 \\
\fontsize{6.5}{6.5}\selectfont 3D-SiamRPN \cite{fang20203d} & \scriptsize 58.2 / 76.2 & \scriptsize 35.2 / 56.2 & \scriptsize 45.7 / 52.9 & \scriptsize 36.2 / 49.0 & \scriptsize 43.8 / 58.6 \\
\fontsize{6.5}{6.5}\selectfont SA-P2B \cite{zhou2021structure} & \scriptsize 58.0 / 75.1 & \scriptsize 34.6 / 63.3 & \scriptsize 51.2 / 63.1 & \scriptsize 32.0 / 43.6 & \scriptsize 44.0 / 61.3 \\
\fontsize{6.5}{6.5}\selectfont MLVSNet \cite{wang2021mlvsnet} & \scriptsize 56.0 / 74.0 & \scriptsize 34.1 / 61.1 & \scriptsize 52.0 / 61.4 & \scriptsize 34.3 / 44.5 & \scriptsize 44.1 / 60.3 \\
\fontsize{6.5}{6.5}\selectfont BAT \cite{zheng2021box} & \scriptsize 60.5 / 77.7 & \scriptsize 42.1 / 70.1 & \scriptsize 52.4 / \textbf{67.0} & \scriptsize 33.7 / 45.4 & \scriptsize 47.2 / 65.1 \\
\fontsize{6.5}{6.5}\selectfont PTT \cite{shan2021ptt} & \scriptsize \textbf{67.8} / \textbf{81.8} & \scriptsize 44.9 / 72.0 & \scriptsize 43.6 / 52.5 & \scriptsize 37.2 / 47.3 & \scriptsize 48.4 / 63.4 \\
\fontsize{6.5}{6.5}\selectfont LTTR \cite{cui20213dlttr} & \scriptsize 65.0 / 77.1 & \scriptsize 33.2 / 56.8 & \scriptsize 35.8 / 45.6 & \scriptsize \textbf{66.2} / 89.9 & \scriptsize 50.0 / 67.4 \\
\hline
\bf \fontsize{6.75}{6.75}\selectfont Ours & \scriptsize 65.2 / 77.4 & \scriptsize \textbf{50.9 / 81.6} & \scriptsize \textbf{52.5} / 61.8 & \scriptsize 65.1 / \textbf{90.5} & \scriptsize \textbf{58.4 / 77.8} \\
\Xhline{1pt}
\end {tabular}}
}
\end{center}
\vspace{-5mm}
\end {table}

\begin{table}[h]
\caption{\textbf{Inference Time.}}
\label{tab:speed_exp} 
\vspace{-7mm}
\begin{center}
\scalebox{0.8}{\begin{tabular}{c|c|c}
\Xhline{1pt}
{\bf SC3D} \cite{2019sc3d} & {\bf P2B } \cite{2020p2b} & {\bf \modelname{} (Ours)} \\ 
\hline\hline
66.3ms  & 23.6ms & \textbf{19.9ms} \\
\Xhline{1pt}
\end {tabular}}
\vspace{-5mm}
\end{center}
\end {table}

\smallskip
\noindent\textbf{Inference Time.} Speed is a key factor in object tracking tasks. Hence, we test the model inference time on the KITTI test dataset with a Tesla V100 GPU. As reported in Tab.~\ref{tab:speed_exp}, under the same configurations, \modelname{} achieves the shortest average runtime of 19.9ms.

\subsection{Ablation Studies} \label{sec:ablations}

To evaluate the effectiveness of the components proposed in \modelname{}, we conduct ablation studies on the KITTI \cite{kitti} dataset and report the Success and Precision.

\begin{table}
\setlength{\tabcolsep}{4.75pt}
\caption {\textbf{Performance comparison on different sampling methods.} D-FPS refers to distance-farthest point sampling and F-FPS denotes feature-farthest point sampling. RAS refers to our proposed Relation-Aware Sampling.} \label{tab:sample_experiment}
\vspace{-6mm}
\begin{center}
\scalebox{0.78}{
\begin{tabular}{l|c|c|c|c|c}
\Xhline{1pt}
{\bf Method} & {\bf Car } & {\bf Pedestrian} & {\bf Van} & {\bf Cyclist} & {\bf Average} \\ 
\hline\hline
Random \cite{2020p2b} & \fontsize{8}{8}\selectfont 62.4 / 74.0 & \fontsize{8}{8}\selectfont 36.6 / 59.9 & \fontsize{8}{8}\selectfont 50.4 / 58.3 & \fontsize{8}{8}\selectfont 62.2 / 83.9 & \fontsize{8}{8}\selectfont 52.9 / 69.0 \\
D-FPS \cite{qi2017pointnet++}& \fontsize{8}{8}\selectfont 61.3 / 73.0 & \fontsize{8}{8}\selectfont 42.5 / 68.8 & \fontsize{8}{8}\selectfont 41.8 / 47.2 & \fontsize{8}{8}\selectfont 59.8 / 78.5 & \fontsize{8}{8}\selectfont 51.4 / 66.9 \\
F-FPS \cite{20203dssd} & \fontsize{8}{8}\selectfont 59.3 / 72.5 & \fontsize{8}{8}\selectfont 41.9 / 68.6 & \fontsize{8}{8}\selectfont 52.1 / 60.5 & \fontsize{8}{8}\selectfont 63.8 / 83.8 & \fontsize{8}{8}\selectfont 54.3 / 71.4 \\
\hline
\bf RAS (Ours) & \fontsize{8.5}{8.5}\selectfont \textbf{65.2 / 77.4} & \fontsize{8.5}{8.5}\selectfont \textbf{50.9 / 81.6} & \fontsize{8.5}{8.5}\selectfont \textbf{52.5 / 61.8} & \fontsize{8.5}{8.5}\selectfont \textbf{65.1 / 90.5} & \fontsize{8.5}{8.5}\selectfont \textbf{58.4 / 77.8}\\
\Xhline{1pt}
\end {tabular}
}
\end{center}
\vspace{-2mm}
\end {table}

\noindent\textbf{Sampling Methods.} 
We compare our proposed Relation-Aware Sampling (RAS) method with existing sampling approaches, including random sampling \cite{2020p2b}, distance-farthest point sampling (D-FPS) \cite{qi2017pointnet++} and feature-farthest point sampling (F-FPS) \cite{20203dssd}. 
As shown in Tab.~\ref{tab:sample_experiment}, RAS yields the best performance with a clear margin. 
By utilizing RAS, our method achieves an increase of 5.5/8.8 in Success/Precision than the random sampling baseline.
Small objects usually consist of fewer points, and hence are more sensitive to the point sparsity challenge. 
For the pedestrian class, which is the class of the smallest object size,
RAS significantly boosts the results from 36.6/59.9 to 50.9/81.6.

\begin {table}[h]
\setlength{\tabcolsep}{6pt}
\begin{center}
\caption{\textbf{Ablation studies on model components.} For experiments that disable the Point Relation Transformer (PRT), we replace PRT with cosine similarity for feature correlation extraction as in existing methods \cite{2020p2b, 2019sc3d, zhou2021structure}. We also compare the performance w/ or w/o the Prediction Refinement Module (PRM).}
\vspace{-2.75mm}
\scalebox{0.675}{
\begin{tabular}{c|c|c|c|c|c|c}
\Xhline{1pt}
{\bf PRT} & {\bf PRM} & {\bf Car } & {\bf Pedestrian} & {\bf Van} & {\bf Cyclist}  & {\bf Average}  \\ \hline\hline
 & & 40.2 / 52.0 & 23.0 / 41.6 & 25.9 / 34.7 & 30.0 / 57.8 & 29.8 / 46.5 \\ 
 \checkmark & & 62.9 / 74.3 & 49.1 / 77.7 & 50.7 / 58.7 & 64.1 / 90.0 & 56.7 / 75.2  \\ 
  & \checkmark  & 60.6 / 73.1 & 39.2 / 66.9 & 43.5 / 48.9 & 58.7 / 87.2 & 50.5 / 69.0\\ 
  \checkmark & \checkmark  & 65.2 / 77.4 & 50.9 / 81.6 & 52.5 / 61.8 & 65.1 / 90.5 & 58.4 / 77.8 \\ 
\Xhline{1pt}
\end {tabular}
}
\label{tab:ablation_components} 
\end{center}
\vspace{-4mm}
\end {table}

\noindent\textbf{Model Components.} 
We conduct experiments to investigate the effectiveness of the proposed Point Relation Transformer (PRT) and Prediction Refinement Module (PRM). 
For the ablation studies on PRT, we replace PRT with cosine similarity for feature correlation computation as in existing methods \cite{2020p2b, zhou2021structure, 2019sc3d}. 
We evaluate with the coarse prediction in experiments w/o PRM. 
As shown in Tab.~\ref{tab:ablation_components}, when both PRT and PRM are disabled, the performance degrades sharply from 58.4 to 29.8 in Success. 
Both PRT and PRM improve the model performance significantly compared to the base case. 
Note that replacing our PRT with cosine similarity will decrease the performance by 8.8 on average in terms of Success, which shows the strong feature matching capability of the PRT module. 
In addition, the two components are also complementary to each other that the best results are obtained when both are enabled. 
We highlight that even without prediction refinement, our model still achieves the best average performance (56.7/75.2).

\vspace{-2mm}
\begin {table}[h]
 \setlength{\tabcolsep}{4.5pt}
\begin{center}
\caption{\textbf{Ablation experiments on Relation Attention.} We investigate the effectiveness of the two major modifications in our proposed Relation Attention module. \textit{Offset} refers to offset attention and \textit{Norm} refers to feature normalization.}
\vspace{-2mm}
\scalebox{0.7}{
\begin{tabular}{c|c|c|c|c|c|c}
\Xhline{1pt}
{\bf Offset} & {\bf Norm} & {\bf Car } & {\bf Pedestrian} & {\bf Van} & {\bf Cyclist} & {\bf Average}  \\ \hline\hline
 & & 55.4 / 68.0 & 36.6 / 65.1 & 34.6 / 38.6 & 55.9 / 78.8 & 45.6 / 62.6 \\ 
 \checkmark & & 56.6 / 69.1 & 40.3 / 67.3 & 48.3 / 59.6 & 63.7 / 90.3 & 52.2 / 71.6 \\ 
  & \checkmark  & 63.7 / 75.3 & 47.1 / 73.5 & 53.0 / 60.4 & 64.1 / 89.5 & 57.0 / 74.7 \\ 
  \checkmark & \checkmark  & 65.2 / 77.4 & 50.9 / 81.6 & 52.5 / 61.8 & 65.1 / 90.5 & 58.4 / 77.8 \\ 
\Xhline{1pt}
\end {tabular}
}
\label{tab:attention_components} 
\end{center}
\vspace{-5mm}
\end {table}

\noindent\textbf{Relation Attention.} The main differences between our proposed Relation Attention and regular transformer attention are the L2-normalization applied on query and key features and the offset attention.
Ablation studies on each component are reported in Tab.~\ref{tab:attention_components}. 
Both two operations
improve the model performance, especially the L2-normalization.
It reveals that the cosine distance facilitates 
point cloud feature matching.

\begin{table}[h]
\setlength{\tabcolsep}{4pt}
\caption {\textbf{Performance comparison on the Waymo SOT Dataset.} Success / Precision are used for evaluation.} \label{tab:waymo_experiment} 
\vspace{-5mm}
\small
\begin{center}
\begin{tabular}{l|c|c|c|c}
\Xhline{1pt}
{\bf Method} & {\bf Vehicle } & {\bf Pedestrian} & {\bf Cyclist} & {\bf Average} \\ 
\hline\hline
\small SC3D \cite{2019sc3d} & \fontsize{8}{8}\selectfont 46.5 / 52.7 & \fontsize{8}{8}\selectfont 26.4 / 37.8 & \fontsize{8}{8}\selectfont 26.5 / 37.6 & \fontsize{8}{8}\selectfont 33.1 / 42.7 \\
\small P2B \cite{2020p2b} & \fontsize{8}{8}\selectfont 55.7 / 62.2 & \fontsize{8}{8}\selectfont 35.3 / 54.9 & \fontsize{8}{8}\selectfont 30.7 / 44.5 & \fontsize{8}{8}\selectfont 40.6 / 53.9 \\
\hline
\bf Ours & \fontsize{8.5}{8.5}\selectfont \textbf{58.7 / 65.2} & \fontsize{8.5}{8.5}\selectfont \textbf{49.0 / 69.1} & \fontsize{8.5}{8.5}\selectfont \textbf{43.3 / 60.4} & \fontsize{8.5}{8.5}\selectfont \textbf{50.3 / 64.9} \\
\Xhline{1pt}
\end {tabular}
\end{center}
\vspace{-4mm}
\end {table}

\subsection{3D Tracking on Waymo SOT Dataset}

For the Waymo SOT Dataset, we compare \modelname{} with SC3D \cite{2019sc3d} and P2B \cite{2020p2b} by re-implementing the methods with the official codes.
As shown in Tab.~\ref{tab:waymo_experiment}, \modelname{} again achieves the best results among all comparing methods with a large margin of 9.7 and 11.0 in average Success and Precision. 
Similar to KITTI, the Pedestrian and Cyclist classes see higher gains.
Overall, the consistent performance improvements on different benchmarks demonstrate the effectiveness and robustness of our proposed method.

\section{Limitation Discussion}
\begin{figure}[htb]
    \centering
    \includegraphics[width=1.0\linewidth]{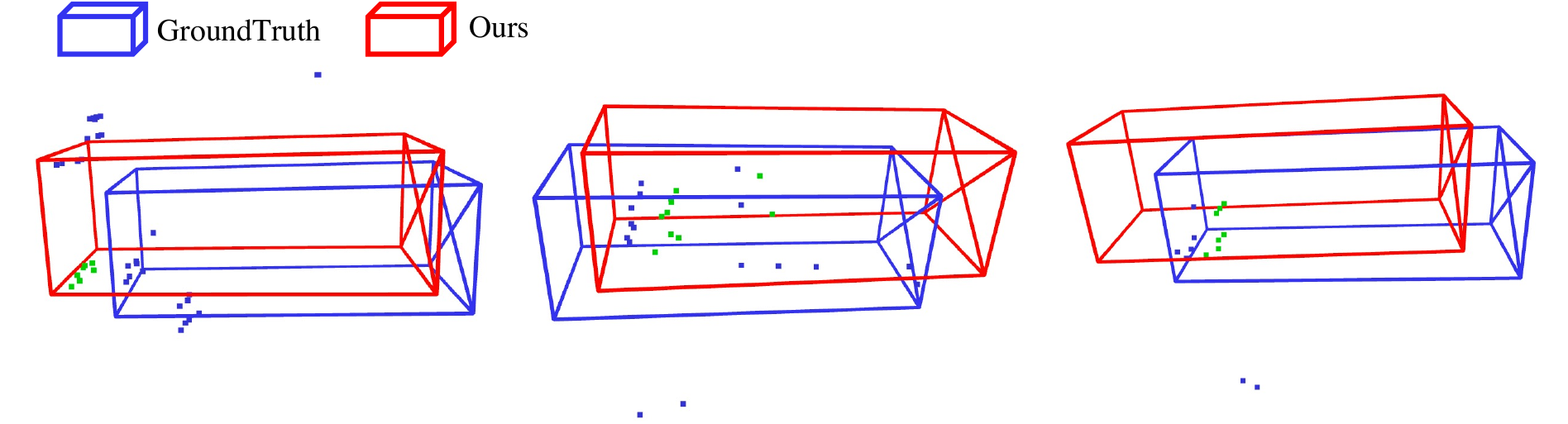}
    \vspace{-8mm}
    \caption{
    Our tracking failures mainly occur when the point clouds are too sparse.}
    \label{fig:failure}
\end{figure}

We show in Fig.~\ref{fig:failure} the failure cases encountered by our model, which mainly occur when the point clouds are too sparse that the model can hardly capture enough patterns to effectively match template and search point clouds. One possible way to further mitigate this issue could be utilizing complementary multi-frame information for object tracking, which can be explored in future researches.

\section{Conclusion}
In this paper, we propose \modelname{}, a novel framework for 3D point cloud single object tracking, which contains a designed Relation-Aware Sampling strategy to tackle point sparsity, a novel Point Relation Transformer for feature matching, and a lightweight Prediction Refinement Module. \modelname{} not only obtains the new state-of-the-art performance but also achieves improved efficiency. We also generate a large-scale SOT tracking dataset based on the Waymo Open Dataset to facilitate more comprehensive evaluations of 3D tracking methods. We hope our method and the Waymo SOT Dataset can help motivate further researches.

\smallskip
\noindent\textbf{Acknowledgement}
This study is supported under the RIE2020 Industry Alignment Fund – Industry Collaboration Projects (IAF-ICP) Funding Initiative, as well as cash and in-kind contribution from the industry partner(s).

\clearpage

{\small
\bibliographystyle{ieee_fullname}
\bibliography{egbib}

\begin{thebibliography}{10}\itemsep=-1pt

\bibitem{agarap2018deep}
Abien~Fred Agarap.
\newblock Deep learning using rectified linear units (relu).
\newblock {\em arXiv preprint arXiv:1803.08375}, 2018.

\bibitem{bertinetto2016fully}
Luca Bertinetto, Jack Valmadre, Joao~F Henriques, Andrea Vedaldi, and Philip~HS
  Torr.
\newblock Fully-convolutional siamese networks for object tracking.
\newblock In {\em European conference on computer vision}, pages 850--865.
  Springer, 2016.

\bibitem{carion2020end}
Nicolas Carion, Francisco Massa, Gabriel Synnaeve, Nicolas Usunier, Alexander
  Kirillov, and Sergey Zagoruyko.
\newblock End-to-end object detection with transformers.
\newblock In {\em European Conference on Computer Vision}, pages 213--229.
  Springer, 2020.

\bibitem{chen2020simple}
Ting Chen, Simon Kornblith, Mohammad Norouzi, and Geoffrey Hinton.
\newblock A simple framework for contrastive learning of visual
  representations.
\newblock In {\em International conference on machine learning}, pages
  1597--1607. PMLR, 2020.

\bibitem{chu2021twins}
Xiangxiang Chu, Zhi Tian, Yuqing Wang, Bo Zhang, Haibing Ren, Xiaolin Wei,
  Huaxia Xia, and Chunhua Shen.
\newblock Twins: Revisiting the design of spatial attention in vision
  transformers.
\newblock {\em arXiv preprint arXiv:2104.13840}, 1(2):3, 2021.

\bibitem{cui20213dlttr}
Yubo Cui, Zheng Fang, Jiayao Shan, Zuoxu Gu, and Sifan Zhou.
\newblock 3d object tracking with transformer.
\newblock {\em arXiv preprint arXiv:2110.14921}, 2021.

\bibitem{engel2021point}
Nico Engel, Vasileios Belagiannis, and Klaus Dietmayer.
\newblock Point transformer.
\newblock {\em IEEE Access}, 9:134826--134840, 2021.

\bibitem{fang20203d}
Zheng Fang, Sifan Zhou, Yubo Cui, and Sebastian Scherer.
\newblock 3d-siamrpn: An end-to-end learning method for real-time 3d single
  object tracking using raw point cloud.
\newblock {\em IEEE Sensors Journal}, 21(4):4995--5011, 2020.

\bibitem{kitti}
Andreas Geiger, Philip Lenz, and Raquel Urtasun.
\newblock Are we ready for autonomous driving? the kitti vision benchmark
  suite.
\newblock In {\em 2012 IEEE conference on computer vision and pattern
  recognition}, pages 3354--3361. IEEE, 2012.

\bibitem{2019sc3d}
Silvio Giancola, Jesus Zarzar, and Bernard Ghanem.
\newblock Leveraging shape completion for 3d siamese tracking.
\newblock In {\em Proceedings of the IEEE/CVF Conference on Computer Vision and
  Pattern Recognition}, pages 1359--1368, 2019.

\bibitem{guo2021pct}
Meng-Hao Guo, Jun-Xiong Cai, Zheng-Ning Liu, Tai-Jiang Mu, Ralph~R Martin, and
  Shi-Min Hu.
\newblock Pct: Point cloud transformer.
\newblock {\em Computational Visual Media}, 7(2):187--199, 2021.

\bibitem{guo2017learning}
Qing Guo, Wei Feng, Ce Zhou, Rui Huang, Liang Wan, and Song Wang.
\newblock Learning dynamic siamese network for visual object tracking.
\newblock In {\em Proceedings of the IEEE international conference on computer
  vision}, pages 1763--1771, 2017.

\bibitem{ioffe2015batch}
Sergey Ioffe and Christian Szegedy.
\newblock Batch normalization: Accelerating deep network training by reducing
  internal covariate shift.
\newblock In {\em International conference on machine learning}, pages
  448--456. PMLR, 2015.

\bibitem{karunasekera2019multiple}
Hasith Karunasekera, Han Wang, and Handuo Zhang.
\newblock Multiple object tracking with attention to appearance, structure,
  motion and size.
\newblock {\em IEEE Access}, 7:104423--104434, 2019.

\bibitem{kingma2014adam}
Diederik~P Kingma and Jimmy Ba.
\newblock Adam: A method for stochastic optimization.
\newblock {\em arXiv preprint arXiv:1412.6980}, 2014.

\bibitem{landrieu2018large}
Loic Landrieu and Martin Simonovsky.
\newblock Large-scale point cloud semantic segmentation with superpoint graphs.
\newblock In {\em Proceedings of the IEEE conference on computer vision and
  pattern recognition}, pages 4558--4567, 2018.

\bibitem{li2019siamrpn++}
Bo Li, Wei Wu, Qiang Wang, Fangyi Zhang, Junliang Xing, and Junjie Yan.
\newblock Siamrpn++: Evolution of siamese visual tracking with very deep
  networks.
\newblock In {\em Proceedings of the IEEE/CVF Conference on Computer Vision and
  Pattern Recognition}, pages 4282--4291, 2019.

\bibitem{li2018high}
Bo Li, Junjie Yan, Wei Wu, Zheng Zhu, and Xiaolin Hu.
\newblock High performance visual tracking with siamese region proposal
  network.
\newblock In {\em Proceedings of the IEEE conference on computer vision and
  pattern recognition}, pages 8971--8980, 2018.

\bibitem{liu2021swin}
Ze Liu, Yutong Lin, Yue Cao, Han Hu, Yixuan Wei, Zheng Zhang, Stephen Lin, and
  Baining Guo.
\newblock Swin transformer: Hierarchical vision transformer using shifted
  windows.
\newblock {\em arXiv preprint arXiv:2103.14030}, 2021.

\bibitem{luo2021unsupervised}
Zhipeng Luo, Zhongang Cai, Changqing Zhou, Gongjie Zhang, Haiyu Zhao, Shuai Yi,
  Shijian Lu, Hongsheng Li, Shanghang Zhang, and Ziwei Liu.
\newblock Unsupervised domain adaptive 3d detection with multi-level
  consistency.
\newblock In {\em Proceedings of the IEEE/CVF International Conference on
  Computer Vision}, pages 8866--8875, 2021.

\bibitem{pan2021variational}
Liang Pan, Xinyi Chen, Zhongang Cai, Junzhe Zhang, Haiyu Zhao, Shuai Yi, and
  Ziwei Liu.
\newblock Variational relational point completion network.
\newblock In {\em Proceedings of the IEEE/CVF Conference on Computer Vision and
  Pattern Recognition}, pages 8524--8533, 2021.

\bibitem{qi2019deep}
Charles~R Qi, Or Litany, Kaiming He, and Leonidas~J Guibas.
\newblock Deep hough voting for 3d object detection in point clouds.
\newblock In {\em Proceedings of the IEEE/CVF International Conference on
  Computer Vision}, pages 9277--9286, 2019.

\bibitem{qi2017pointnet}
Charles~R Qi, Hao Su, Kaichun Mo, and Leonidas~J Guibas.
\newblock Pointnet: Deep learning on point sets for 3d classification and
  segmentation.
\newblock In {\em Proceedings of the IEEE conference on computer vision and
  pattern recognition}, pages 652--660, 2017.

\bibitem{qi2017pointnet++}
Charles~R Qi, Li Yi, Hao Su, and Leonidas~J Guibas.
\newblock Pointnet++: Deep hierarchical feature learning on point sets in a
  metric space.
\newblock {\em arXiv preprint arXiv:1706.02413}, 2017.

\bibitem{2020p2b}
Haozhe Qi, Chen Feng, Zhiguo Cao, Feng Zhao, and Yang Xiao.
\newblock P2b: Point-to-box network for 3d object tracking in point clouds.
\newblock In {\em Proceedings of the IEEE/CVF Conference on Computer Vision and
  Pattern Recognition}, pages 6329--6338, 2020.

\bibitem{ren2022benchmarking}
Jiawei Ren, Liang Pan, and Ziwei Liu.
\newblock Benchmarking and analyzing point cloud classification under
  corruptions.
\newblock {\em arXiv preprint arXiv:2202.03377}, 2022.

\bibitem{ren2015faster}
Shaoqing Ren, Kaiming He, Ross Girshick, and Jian Sun.
\newblock Faster r-cnn: Towards real-time object detection with region proposal
  networks.
\newblock {\em Advances in neural information processing systems}, 28:91--99,
  2015.

\bibitem{shan2021ptt}
Jiayao Shan, Sifan Zhou, Zheng Fang, and Yubo Cui.
\newblock Ptt: Point-track-transformer module for 3d single object tracking in
  point clouds.
\newblock In {\em 2021 IEEE/RSJ International Conference on Intelligent Robots
  and Systems (IROS)}, pages 1310--1316. IEEE, 2021.

\bibitem{shi2019pointrcnn}
Shaoshuai Shi, Xiaogang Wang, and Hongsheng Li.
\newblock Pointrcnn: 3d object proposal generation and detection from point
  cloud.
\newblock In {\em Proceedings of the IEEE/CVF conference on computer vision and
  pattern recognition}, pages 770--779, 2019.

\bibitem{waymo}
Pei Sun, Henrik Kretzschmar, Xerxes Dotiwalla, Aurelien Chouard, Vijaysai
  Patnaik, Paul Tsui, James Guo, Yin Zhou, Yuning Chai, Benjamin Caine, et~al.
\newblock Scalability in perception for autonomous driving: Waymo open dataset.
\newblock In {\em Proceedings of the IEEE/CVF Conference on Computer Vision and
  Pattern Recognition}, pages 2446--2454, 2020.

\bibitem{tao2016siamese}
Ran Tao, Efstratios Gavves, and Arnold~WM Smeulders.
\newblock Siamese instance search for tracking.
\newblock In {\em Proceedings of the IEEE conference on computer vision and
  pattern recognition}, pages 1420--1429, 2016.

\bibitem{vaswani2017attention}
Ashish Vaswani, Noam Shazeer, Niki Parmar, Jakob Uszkoreit, Llion Jones,
  Aidan~N Gomez, {\L}ukasz Kaiser, and Illia Polosukhin.
\newblock Attention is all you need.
\newblock In {\em Advances in neural information processing systems}, pages
  5998--6008, 2017.

\bibitem{wang2017normface}
Feng Wang, Xiang Xiang, Jian Cheng, and Alan~Loddon Yuille.
\newblock Normface: L2 hypersphere embedding for face verification.
\newblock In {\em Proceedings of the 25th ACM international conference on
  Multimedia}, pages 1041--1049, 2017.

\bibitem{wang2018cosface}
Hao Wang, Yitong Wang, Zheng Zhou, Xing Ji, Dihong Gong, Jingchao Zhou, Zhifeng
  Li, and Wei Liu.
\newblock Cosface: Large margin cosine loss for deep face recognition.
\newblock In {\em Proceedings of the IEEE conference on computer vision and
  pattern recognition}, pages 5265--5274, 2018.

\bibitem{wang2021transformer}
Ning Wang, Wengang Zhou, Jie Wang, and Houqiang Li.
\newblock Transformer meets tracker: Exploiting temporal context for robust
  visual tracking.
\newblock In {\em Proceedings of the IEEE/CVF Conference on Computer Vision and
  Pattern Recognition}, pages 1571--1580, 2021.

\bibitem{wang2021joint}
Yongxin Wang, Kris Kitani, and Xinshuo Weng.
\newblock Joint object detection and multi-object tracking with graph neural
  networks.
\newblock In {\em 2021 IEEE International Conference on Robotics and Automation
  (ICRA)}, pages 13708--13715. IEEE, 2021.

\bibitem{wang2021mlvsnet}
Zhoutao Wang, Qian Xie, Yu-Kun Lai, Jing Wu, Kun Long, and Jun Wang.
\newblock Mlvsnet: Multi-level voting siamese network for 3d visual tracking.
\newblock In {\em Proceedings of the IEEE/CVF International Conference on
  Computer Vision}, pages 3101--3110, 2021.

\bibitem{weng2019baseline}
Xinshuo Weng and Kris Kitani.
\newblock A baseline for 3d multi-object tracking.
\newblock {\em arXiv preprint arXiv:1907.03961}, 1(2):6, 2019.

\bibitem{weng20203d}
Xinshuo Weng, Jianren Wang, David Held, and Kris Kitani.
\newblock 3d multi-object tracking: A baseline and new evaluation metrics.
\newblock In {\em 2020 IEEE/RSJ International Conference on Intelligent Robots
  and Systems (IROS)}, pages 10359--10366. IEEE, 2020.

\bibitem{weng2020graph}
Xinshuo Weng, Yongxin Wang, Yunze Man, and Kris Kitani.
\newblock Graph neural networks for 3d multi-object tracking.
\newblock {\em arXiv preprint arXiv:2008.09506}, 2020.

\bibitem{xiang2015learning}
Yu Xiang, Alexandre Alahi, and Silvio Savarese.
\newblock Learning to track: Online multi-object tracking by decision making.
\newblock In {\em Proceedings of the IEEE international conference on computer
  vision}, pages 4705--4713, 2015.

\bibitem{xiao2022unsupervised}
Aoran Xiao, Jiaxing Huang, Dayan Guan, and Shijian Lu.
\newblock Unsupervised representation learning for point clouds: A survey.
\newblock {\em arXiv preprint arXiv:2202.13589}, 2022.

\bibitem{xiao2021synlidar}
Aoran Xiao, Jiaxing Huang, Dayan Guan, Fangneng Zhan, and Shijian Lu.
\newblock Synlidar: Learning from synthetic lidar sequential point cloud for
  semantic segmentation.
\newblock {\em arXiv preprint arXiv:2107.05399}, 2021.

\bibitem{xie2021segformer}
Enze Xie, Wenhai Wang, Zhiding Yu, Anima Anandkumar, Jose~M Alvarez, and Ping
  Luo.
\newblock Segformer: Simple and efficient design for semantic segmentation with
  transformers.
\newblock {\em arXiv preprint arXiv:2105.15203}, 2021.

\bibitem{20203dssd}
Zetong Yang, Yanan Sun, Shu Liu, and Jiaya Jia.
\newblock 3dssd: Point-based 3d single stage object detector.
\newblock In {\em Proceedings of the IEEE/CVF conference on computer vision and
  pattern recognition}, pages 11040--11048, 2020.

\bibitem{yin2021center}
Tianwei Yin, Xingyi Zhou, and Philipp Krahenbuhl.
\newblock Center-based 3d object detection and tracking.
\newblock In {\em Proceedings of the IEEE/CVF Conference on Computer Vision and
  Pattern Recognition}, pages 11784--11793, 2021.

\bibitem{zhang2021meta}
Gongjie Zhang, Zhipeng Luo, Kaiwen Cui, and Shijian Lu.
\newblock Meta-detr: Few-shot object detection via unified image-level
  meta-learning.
\newblock {\em arXiv preprint arXiv:2103.11731}, 2(6), 2021.

\bibitem{zhang2022accelerating}
Gongjie Zhang, Zhipeng Luo, Yingchen Yu, Kaiwen Cui, and Shijian Lu.
\newblock Accelerating detr convergence via semantic-aligned matching.
\newblock {\em arXiv preprint arXiv:2203.06883}, 2022.

\bibitem{zhao2020exploring}
Hengshuang Zhao, Jiaya Jia, and Vladlen Koltun.
\newblock Exploring self-attention for image recognition.
\newblock In {\em Proceedings of the IEEE/CVF Conference on Computer Vision and
  Pattern Recognition}, pages 10076--10085, 2020.

\bibitem{zhao2021point}
Hengshuang Zhao, Li Jiang, Jiaya Jia, Philip~HS Torr, and Vladlen Koltun.
\newblock Point transformer.
\newblock In {\em Proceedings of the IEEE/CVF International Conference on
  Computer Vision}, pages 16259--16268, 2021.

\bibitem{zheng2021box}
Chaoda Zheng, Xu Yan, Jiantao Gao, Weibing Zhao, Wei Zhang, Zhen Li, and
  Shuguang Cui.
\newblock Box-aware feature enhancement for single object tracking on point
  clouds.
\newblock In {\em Proceedings of the IEEE/CVF International Conference on
  Computer Vision}, pages 13199--13208, 2021.

\bibitem{zhou2021structure}
Xiaoyu Zhou, Ling Wang, Zhian Yuan, Ke Xu, and Yanxin Ma.
\newblock Structure aware 3d single object tracking of point cloud.
\newblock {\em Journal of Electronic Imaging}, 30(4):043010, 2021.

\bibitem{zhu2020deformable}
Xizhou Zhu, Weijie Su, Lewei Lu, Bin Li, Xiaogang Wang, and Jifeng Dai.
\newblock Deformable detr: Deformable transformers for end-to-end object
  detection.
\newblock {\em arXiv preprint arXiv:2010.04159}, 2020.

\end{thebibliography}
}

\end{document}